\documentclass[sigconf]{acmart}
\AtBeginDocument{%
  \providecommand\BibTeX{{%
    \normalfont B\kern-0.5em{\scshape i\kern-0.25em b}\kern-0.8em\TeX}}}

\copyrightyear{2021}
\acmYear{2021}
\setcopyright{acmcopyright}\acmConference[MM '21]{Proceedings of the 29th ACM International Conference on Multimedia}{October 20--24, 2021}{Virtual Event, China}
\acmBooktitle{Proceedings of the 29th ACM International Conference on Multimedia (MM '21), October 20--24, 2021, Virtual Event, China}
\acmPrice{15.00}
\acmDOI{10.1145/3474085.3479231}
\acmISBN{978-1-4503-8651-7/21/10}

\usepackage{multirow}
\newcommand{\ie}{\textit{i}.\textit{e}.}

\usepackage{pdfrender}
\newcommand*{\boldcheckmark}{%
  \textpdfrender{
    TextRenderingMode=FillStroke,
    LineWidth=.5pt, % half of the line width is outside the normal glyph
  }{\checkmark}%
}
\usepackage{arydshln} % for \hdashline
\usepackage{enumitem} % for indent
\usepackage{bm}

\definecolor{green}{RGB}{24, 181, 0}

\newcommand\blfootnote[1]{%
  \begingroup
  \renewcommand\thefootnote{}\footnote{#1}%
  \addtocounter{footnote}{-1}%
  \endgroup
}
\begin{document}
\fancyhead{}

% \title{TempFormer: Temporal-aware Transformer for Video Scene Graph Generation}
\title{Video Relation Detection via Tracklet based Visual Transformer}

\author{
   Kaifeng Gao$^\dagger$,
   Long Chen$^{\ddagger*}$,
   Yifeng Huang$^\dagger$, and
   Jun Xiao$^\dagger$
}
\affiliation{%
  \institution{$^\dagger$Zhejiang University \qquad $^\ddagger$Columbia University}
%   \institution{\textsuperscript{\rm 2} Columbia University, New York, USA}
  \city{}
  \country{}
}
\email{kite_phone@zju.edu.cn, zjuchenlong@gmail.com, yfhuang@zju.edu.cn, junx@cs.zju.edu.cn}


\begin{abstract}
\textbf{Vid}eo \textbf{V}isual \textbf{R}elation \textbf{D}etection (VidVRD), has received significant attention of our community over recent years. In this paper, we apply the state-of-the-art video object tracklet detection pipeline MEGA~\cite{chen2020memory} and deepSORT~\cite{wojke2017simple} to generate tracklet proposals. Then we perform VidVRD in a tracklet-based manner without any pre-cutting operations. Specifically, we design a tracklet-based visual Transformer. It contains a temporal-aware decoder which performs feature interactions between the tracklets and learnable predicate query embeddings, and finally predicts the relations. Experimental results strongly demonstrate the superiority of our method, which outperforms other methods by a large margin on the Video Relation Understanding (VRU) Grand Challenge in ACM Multimedia 2021. Codes are released at \href{https://github.com/Dawn-LX/VidVRD-tracklets}{https://github.com/Dawn-LX/VidVRD-tracklets}. \blfootnote{$^*$Long Chen is the corresponding author. This work started when Long Chen at ZJU.}
\end{abstract}

   




\maketitle

\section{Introduction}

The Video Visual Relation Detection (VidVRD) task aims to detect visual relations between objects in videos, which are denoted by a set of <\texttt{subject}, \texttt{predicate}, \texttt{object}> triplets. Compared to visual relation detection in still images (ImgVRD)~\cite{chen2019counterfactual}, VidVRD is technically more challenging: (1) The relationship between objects incorporates temporal informations. Some relationships (e.g., \texttt{towards}, \texttt{move-past}) can only be detected by utilizing temporal context. (2) Relations between two specific objects often changes overtime.

Inspired from ImgVRD approaches, early VidVRD models~\cite{shang2017video,qian2019video,tsai2019video,su2020video} are all \textbf{segment-based} approaches. Specifically, they first divide the whole video into multiple short segments. Then, they detect all tracklets and their pairwise relationships in each short segment. Lastly, they merge relationship triplets among adjacent segments with association methods. Although these segment-based VidVRD approaches achive sound performance on standard benchmarks, this framework inherently fails to utilize the long-term temporal context in other segments. 
% And they substantially degrade the VidVRD task into ImgVRD task, \ie, following the ImgSGG framework except replacing the object features with segment-level tracklet features. 
To avoid this inherent limitation, one of the latest VidVRD model STGCN~\cite{liu2020beyond} performs VidVRD in a \textbf{tracklet-based} manner. Instead of pre-cutting the input video into multiple segments in advance, STGCN directly detects all object tracklets, and runs sliding windows with different scales to obtain numerous tracklet proposals. Then, STGCN predicts the tracklet classes and their pairwise relationships.

In this paper, we follow STGCN~\cite{liu2020beyond} to first detect all the object tracklets in a video. Then we design a tracklet based visual Transformer to perform interactions between tracklet features and finally detect the relations. Specifically, Our model is a Transformer-family encoder-decoder model~\cite{vaswani2017attention}, where the inputs for the encoder and decoder are initial tracklet features and learnable predicate query embeddings, respectively. The encoder aims to extract contextual features for object tracklets, and the decoder intends to enhance predicate queries with all tracklets. We use each predicate query to represent a relationship instance, and it has two types (\ie, \texttt{subject}/\texttt{object}) of links to tracklets. The final output of our model are the enhanced predicate queries, which will be fed into a multi-layer perceptron (MLP) for predicate classification, and the attention matrix from the last encoder layer, which will be binarized to obtain the links to \texttt{subject}/\texttt{object} for each predicate query.

We will describe our method in two main steps: 1) object detection and tracking (Section~\ref{sec_2}), which returns a set of object tracklet proposals, and 2) relation detection (Section~\ref{sec_3}), which detects the visual relations among the pre-obtained tracklet proposals.

\section{Object Detection and tracking}\label{sec_2}
\begin{figure*}[t]
   \begin{center}
      \includegraphics[width=\linewidth]{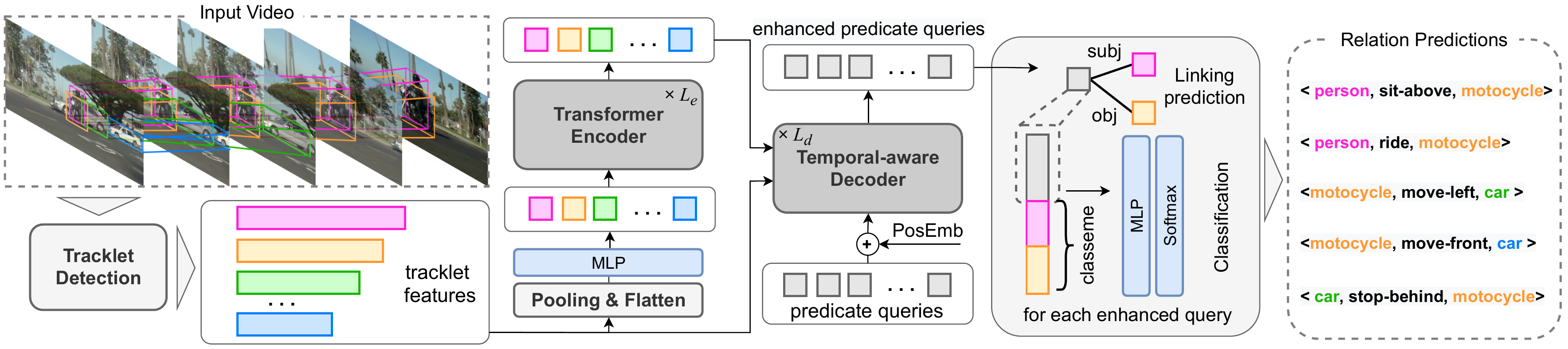}
   \end{center}
   \vspace{-0.5em}
   \caption{The pipeline of our method. Note that each predicate query is assigned with a pre-defined temporal anchor, and the positional embeddings (PosEmb) are calculated based on these anchors through a learnable projection matrix.
   }
   \label{fig:pipeline}
\end{figure*}

Object tracklet detection determines the ceiling of the video visual relation detection performance. Different from previous works, we use MEGA~\cite{chen2020memory} with ResNet-101~\cite{he2016deep} to detect frame-level objects, and then use deepSORT~\cite{wojke2017simple} to associate them into tracklet proposals.

Our detector was trained on the training and validation set of MS-COCO\cite{lin2014microsoft} and the training set of VidOR~\cite{shang2019annotating}. For MS-COCO, we selected the same 80 object categories as that in VidOR. For VidOR, considering the redundancy of adjacent frames, 
% due to the large redundancy between adjacent frames of videos, 
we sample key frames every 32 frames for each video. After all, the training set consists around 311k images.
%  (84k from MS-COCO and 227k from VidOR) images. 
% Finally this trained detector achieve a frame-level mAP(IoU=0.5) of 41.21\% on the VidOR validation set.

We run the above trained detector on each frame of the video to obtain the frame-level object detection results, where each result contains box coordinates, visual appearance features, and the object classification logits. Then we adopt the object tracking method deepSORT~\cite{wojke2017simple} to generate object tracklets, denoted as $\{T_i\}_{i=1}^n$. Each tracklet $T_i$ with length (number of frames) $l_i$ is characterized by its time slots $(s_i,e_i)$, box coordinates $\{\bm{b}_{i,j} \in \mathbb{R}^{4}~|~j=1,2,\ldots l_i\}$, appearance features $\bm{f}^a_i \in \mathbb{R}^{l_i \times d_a}$, and the object category $c_i \in \mathcal{C}_\text{obj}$, where $\mathcal{C}_\text{obj}$ is the set of all object categories for VidOR. We fix all the box coordinates $\{\bm{b}_{i,j}\}$ and object categories $\{c_i\}$ as the final predictions. In addition, we reserve the classification probabilities from the detection backbone for each tracklet, denoted as $\bm{p}_i \in \mathbb{R}^{|\mathcal{C}_\text{obj}|}$, for the final relation prediction.

\section{relation detection}\label{sec_3}
% \gkf{overview}
% Suppose we obtain $n$ tracklet proposals from the detection stage. \gkf{mention n,m, fixed m}
The overall pipeline of our model is shown in Figure~\ref{fig:pipeline}. It uses a fixed set of $m$ predicate queries $\{q_j\}_{j=1}^m$, with learnable query embeddings, denoted as $\bm{Q} \in \mathbb{R}^{m \times d_q}$. Each query is responsible for each of the final predicate prediction, which is characterized by its links to subject/object and its category $c_j^q \in \mathcal{C}_\text{rel}$, where $\mathcal{C}_\text{rel}$ is the set of all relation categories for VidOR.

\subsection{Tracklet Feature Initialization}
For each tracklet $T_i$, we consider both the static coordinates $\{\bm{b}_{i,j}\}$ and the dynamic features $\{\Delta \bm{b}_{i,j}\}$ for its spatial feature, where%\footnote{we set the last one $\Delta \bm{b}_{i,l_i} = \Delta \bm{b}_{i,l_i-1}$ to pad $\bm{s}_i$.}
\begin{align}
   \Delta \bm{b}_{i,j} = \bm{b}_{i,j+1} - \bm{b}_{i,j},~ j=1,\ldots,l_i-1.
\end{align}
The spatial feature $\bm{s}_i \in \mathbb{R}^{l_i \times 8}$ is obtained by stacking $\{\bm{b}_{i,j}\}_{j=1}^{l_i}$ and $\{\Delta \bm{b}_{i,j}\}_{j=1}^{l_i}$. Then, the appearance feature $\bm{f}^a_i$ and spatial feature $\bm{s}_i$ are fed into two MLPs and their outputs are concatenated as the initial tracklet feature $\bm{f}_i \in \mathbb{R}^{l_i\times d}$, where $\bm{f}_i = [\text{MLP}_v(\bm{f}^a_i);\text{MLP}_s(\bm{s}_i)]$.
% \begin{align}
%    \bm{f}_i = [\text{MLP}_v(\bm{f}^a_i);\text{MLP}_s(\bm{s}_i)].
% \end{align}

\subsection{Encoder-Decoder Feature Interactions}
\noindent\textbf{Encoder.} Since the size of each tracklet feature is different, we first use a pooling operation to transform tracklet feature $\bm{f}_i \in \mathbb{R}^{l_i \times d}$ to a fixed size feature $\bm{f}_i \in \mathbb{R}^{l \times d}$, and flatten it into a vector, followed by a MLP to reduce the dimension, resulting in a vector $\bm{h}_{i} \in \mathbb{R}^{d}$. Then, we stack all tracklet features $\{\bm{h}_{i}\}$ into matrix $\bm{H} \in \mathbb{R}^{n \times d}$, and feed $\bm{H}$ into a Transformer encoder, \ie, $\widetilde{\bm{H}} = \text{Transformer}_\text{enc}\left(\bm{H}\right)$, where the outputs $\widetilde{\bm{H}} \in \mathbb{R}^{n \times d}$  are contextualized tracklet features 

\noindent\textbf{Decoder.} In contrast to the standard Transformer decoder, we design a temporal-aware decoder that considers different temporal RoIs of the tracklet features $\{\bm{f}_i\}$. To naturally differentiate each query embedding, we assign each query $q_j$ with a temporal anchor $(s_j^a,e_j^a)$. The time slot of $q_j$ is regressed with respect to $(s_j^a,e_j^a)$ in each decoder layer, denoted as $(s_j^q,e_j^q)$, based on which we perform RoI pooling for each tracklet-query pair:
\begin{align}
   \bm{f}_{i,j}^{\rm roi} & = \text{RoIPool} \left( \bm{f}_{i},(s_{i,j},e_{i,j}) \right ), \quad \text{s.t.}~ (s_{i,j}, e_{i,j}) \neq \emptyset. \\
   \text{where} \quad & (s_{i,j},e_{i,j}) = (s_i,e_i)\cap (s^q_j,e^q_j), \notag
\end{align}
where RoIPool is a one-dim RoI pooling operation, $\bm{f}_{i,j}^{\rm roi} \in \mathbb{R}^{l_\text{roi}\times d}$ is the RoI feature of tracklet $T_i$ for the query $q_j$. Zero-padding is utilized for those tracklet-predicate pairs that have no temporal overlap. After obtaining $\{\bm{f}_{i,j}^{\rm roi}\}$, we flatten them and use a MLP to reduce the dimension:
\begin{align}
   \bm{v}_{i,j} = \text{MLP}_\text{roi}(\text{Flatten}(\bm{f}_{i,j}^{\rm roi})) \in \mathbb{R}^{d_v}.
\end{align}
We stack $\{\bm{v}_{i,j}\}_{i=1}^{n}$ into $\bm{V}_j \in \mathbb{R}^{n\times d_v}$, corresponding to the value matrix of the cross-attention operation in the Transformer decoder. 

However, in contrast to the vanilla cross-attention where a single value matrix is used, our temporal-aware cross-attention 1) constructs separate value matrix $\bm{V}_j$ for each query $q_j$, which considers different temporal information of tracklet-query pairs, and 2) designs role-specific attention matrices for \texttt{subject} ($\bm{A}_s$) and \texttt{object} ($\bm{A}_o$), respectively. Specifically, 
\begin{align}
   \bm{A}_r = 1/\sqrt{d} (\widetilde{\bm{Q}}\bm{W}_r^Q)(\widetilde{\bm{H}}\bm{W}_r^K)^{\rm T},
   \label{eq:cal_Ar}
\end{align}
where $r\in \{s,o\}$ represents semantic roles, $\widetilde{\bm{Q}}$ is the query matrix after self-attention, $\widetilde{\bm{H}}$ is the output of the encoder (served as the key matrix), and $\bm{W}_r^Q$, $\bm{W}_r^K$ are learnable weights. In our setting, we assume each predicate query can only link to one tracklet in each role and each tracklet-predicate pair has one type of roles at most. Thus, we stack $\bm{A}_\text{s}$ and $\bm{A}_\text{o}$ as $\bm{A}\in\mathbb{R}^{2\times m \times n}$, and normalize it through:
\begin{align}
   \widetilde{\bm{A}}[r,j,i] = \frac{\exp(\bm{A}[r,j,i])}{\sum_{i'=1}^{n} \exp(\bm{A}[r,j,i'])} 
   \times \frac{\exp(\bm{A}[r,j,i])}{\sum_{r'=1}^{2} \exp(\bm{A}[r',j,i])}.\label{doubel_softmax_A}
\end{align}
Thereafter, our temporal-aware cross-attention is performed as
\begin{align}\label{eq_Fr}
   \widetilde{\bm{q}}_j = \textstyle{\sum}_{r=1}^2 F_r(\widetilde{\bm{A}}[r,j,:]\bm{V}_j) \in \mathbb{R}^{1\times d_q},~j=1,\ldots,m,
\end{align}
where $\widetilde{\bm{A}}[r,j,:]\in \mathbb{R}^{1 \times n}$ and $F_r: \mathbb{R}^{d_v}\rightarrow \mathbb{R}^{d_q}$ is a role-specific MLP that introduces role-wise distinction into $\widetilde{\bm{q}}_j$.

Finally, we stack $\{\widetilde{\bm{q}}_j\}_{j=1}^{m}$ as the enhanced query embedding matrix of shape $m\times d_q$. The enhanced query embeddings are further utilized to regress the offset w.r.t temporal anchors through an extract MLP, which will be used in the next decoder layer. Other operations such as feed-forward network (FFN), residual connections and layer normalization~\cite{ba2016layer} are also used in our temporal-aware decoder as that in the standard Transformer decoder.

\subsection{Relation Prediction}
Given the attention matrix $\widetilde{\bm{A}} \in \mathbb{R}^{2\times m \times n}$ and the enhanced predicate queries $\{\widetilde{\bm{q}}_j\}_{j=1}^{m}$ output from the last decoder layer, we perform relation prediction through two steps of linking prediction and predicate classification.

For linking prediction, we binarize $\widetilde{\bm{A}}$ to obtain the links of each role (subject/object). Specifically, for each query in each channel (\ie, each $\widetilde{\bm{A}}[r,j,:]$), the tracklet of the max attention score is selected.

For predicate classification, we use the enhanced query embedding $\widetilde{\bm{q}}_j$, the subject/object classeme features, and the frequency bias (\ie, prior information) in the VidOR training set.

Specifically, we use GloVe embeddings~\cite{pennington2014glove} to represent the words of object categories, and stack them as the embedding matrix $\bm{E}\in \mathbb{R}^{|\mathcal{C}_\text{obj}| \times d_w}$, where $d_w$ is the dimension of word embedding. The classeme feature for each tracklet proposal is calculated as $\bm{f}^c_i = \bm{E}^{\rm T}\bm{p}_i \in \mathbb{R}^{d_w}$. Then, we concatenate the predicate query with the classeme features of its corresponding subject-object pair, denoted as $\widetilde{\bm{q}}'_j = [\widetilde{\bm{q}}_j;\bm{f}^c_{i_{j,s}};\bm{f}^c_{i_{j,o}}] \in \mathbb{R}^{d_q + 2d_w}$, where $i_{j,s},i_{j,o}$ index the subject and object tracklets for query $q_j$, respectively.

For frequency bias, we construct a dictionary $\bm{B}$ to store the log-probability of predicate category for a given category pair of subject-object pair~\cite{zellers2018neural}, where $\bm{B}:\mathcal{C}_\text{obj}\times \mathcal{C}_\text{obj} \rightarrow \mathbb{R}^{|\mathcal{C}_\text{rel}|}$. Finally, the predicate classification is performed as 
\begin{align}
   p_j(c^q) \sim \text{softmax}(\text{MLP}_q(\widetilde{\bm{q}}'_j) + \bm{B}[c_{i_{j,s}},c_{i_{j,s}}]),
\end{align}
where $c_{i_{j,s}}$ ($c_{i_{j,o}}$) is the object category of $i_{j,s}$-th ($i_{j,o}$-th) tracklet, and $p_j(c^q)$ is the probability of predicate category $c^{q}$.
\subsection{Training Objectives}
% Our method infers a fixed-size set of $m$ relation predictions. 
Because we fix all the tracklet proposals from the detection backbone as the final predictions, we only consider the training loss between predicates and their ground-truths. Let us denote by $\hat{q}=\{\hat{q}_j\}_{j=1}^m$ the set of $m$ predicate predictions. Let $q^*$ be the ground-truth predicate set of size $m$ padded with $\varnothing$ (background). We adopt one-to-one label assignment by finding a bipartite matching between $\hat{q}$ and $q^*$. Specifically, we search for a permutation of $m$ elements $\hat{\sigma}$ by optimizing the following cost:
\begin{align} %\mathop{\arg\min}_{\sigma}
   \hat{\sigma} = \mathop{\arg\min}_{\sigma} \textstyle{\sum}_{j=1}^{m} \mathcal{L}_{\text{match}}( q^*_j,\hat{q}_{\sigma(j)}).
\end{align}
This matching problem is computed efficiently with the Hungarian algorithm~\cite{munkres1957algorithms}, following prior work~\cite{carion2020end}. The matching cost considers both predicate classification and linking prediction. Because each predicate is characterized by its category and two links, we denote $q^*_j = (c^{q*}_j,\bm{a}^*_j)$, where $c^{q*}_j$ is the predicate category (which may be $\varnothing$) and $\bm{a}^*_j \in \{0,1\}^{2\times n}$ is the $j$-th row of $\bm{A}^*$ (binarized ground-truth attention matrix) for two channels. Note that $\bm{a}^*_j[r,i]=0$ when the $i$-th tracklet has no ground-truth to match (tracklet assignment is based on vIoU and the criterion is similar to that in Faster-RCNN~\cite{ren2015faster}). For the predicted predicate with index $\sigma(j)$, we define the linking prediction as $\hat{\bm{a}}_{\sigma(j)} \in \mathbb{R}^{2\times n}$. With the above notations, the matching cost is defined as
\begin{align}
   \mathcal{L}_{\text{match}}( q^*_j,\hat{q}_{\sigma(j)})= -\mathbf{1}_{\{c^{q*}_j\neq \varnothing\}} \lambda_\text{cls} \log p_{\sigma(j)}(c^{q*}_j) \\ \notag
   + \mathbf{1}_{\{c^{q*}_j\neq \varnothing\}} \lambda_\text{att} \mathcal{L}_\text{att}(\bm{a}^*_j,\hat{\bm{a}}_{\sigma(j)}),
\end{align}
where $\lambda_\text{att}$ and $\lambda_\text{att}$ are hyperparameters, and $\mathcal{L}_\text{att}$ is defined as a binary-cross entropy (BCE) loss.

After obtaining $\hat{\sigma}$, the final predicate loss $\mathcal{L}$ consists of two parts: the matching loss between the matched <$q^*_j,\hat{q}_{\hat{\sigma}(j)}$> pairs, and the background classification loss for other predicate predictions, \ie,
\begin{align}
    \mathcal{L} =  \textstyle{\sum}_{j=1}^{m} \mathcal{L}_{\text{match}}\left( q^*_j,\hat{q}_{\sigma(j)}\right) - \lambda_\text{cls}  \textstyle{\sum}_{c^{q*}_j= \varnothing}\log p_{\sigma(j)}(\varnothing).
\end{align}

\section{Experiments}

\addtolength{\tabcolsep}{-1pt}
\begin{table*}[t]
   \begin{center}
      \begin{tabular}{cc|ccc|ccc|ccc}
         \hline
         \multirow{2}{*}{Team name} & \multirow{2}{*}{Grade} & \multirow{2}{*}{Detector} & \multirow{2}{*}{Tracker} & \multirow{2}{*}{Features} & \multicolumn{3}{c|}{RelDet} & \multicolumn{3}{c}{RelTag} \\
                                    &                        &                           &                          &                           & mAP     & R@50    & R@100   & P@1     & P@5     & P@10   \\ \hline
         RELAbuilder~\cite{zheng2019relation}                & 2nd-2019               & VGG-16 + R-Det           & GTG + T-NMS              & I3D+RM+L                & 0.546   & ---       & ---       & ---       & 23.60   & ---      \\
         MAGUS.Gamma~\cite{sun2019video}    & 1st-2019               & FGFA                      & Seq-NMS + KCF            & RM+L                    & 6.31    & ---       & ---       & ---       & 42.10   & ---      \\
         ETRI\_DGRC                 & 2nd-2020               & ---                         & ---                        & ---                         & 6.65    & ---       & ---       & ---      & ---       & ---      \\
         colab-BUAA~\cite{xie2020video}  & 1st-2020               & S-101 + C-RCNN            & iSeq-NMS                 & V+L+RM+Msk            & 11.74   & 10.02   & 12.69   & 71.36   & 56.30   & 44.59  \\
         EgoJ                       & 3rd-2021               & ---                         & ---                        & ---                         & 5.93    & ---       & ---       & ---       & ---       & ---      \\
         Planck                     & 2nd-2021               & ---                         & ---                        & ---                         & 6.69    & ---       & ---       & ---       & ---       & ---      \\ \hline
         Ours (Ens-5)               & 1st-2021               & R-101 + MEGA              & deepSORT                 & V+L                   & 9.48    & 8.56    & 10.43   & 63.46   & 54.07   & 41.94  \\ \hline
      \end{tabular}
   \end{center}
   \caption{Performance (\%) on VidOR-test (VRU Challenges) of SOTA methods. We list all of their detector/tracker/features if available. \emph{I3D} denotes the I3D~\cite{carreira2017quo} features of object tracklets. \emph{RM} denotes the pair-wise relative motion features of subject-object tracklet pairs. \emph{Msk} is the location mask feature. \emph{V} and \emph{L} are visual and language features of tracklets, respectively.
   }
   \vspace{-1.5em}
   \label{table_test_map}
\end{table*}
\addtolength{\tabcolsep}{1pt}

\addtolength{\tabcolsep}{-0.5pt}
\begin{table}[t]
   \begin{center}
      \begin{tabular}{cc|cc|cc}
         \hline
         Category      & mAP    & Category      & mAP    & Category      & mAP    \\ \hline
         turtle     & 95.61 & sheep/goat & 55.53 & faucet     & 26.87 \\
         adult      & 84.51 & kangaroo   & 53.26 & scooter    & 17.75 \\
         bird       & 81.81 & bus/truck  & 51.49 & cellphone  & 14.77 \\
         motorcycle & 75.13 & watercraft & 47.62 & cattle/cow & 7.11 \\
         baby       & 66.98 & vegetables & 43.13 & fruits     & 1.81 \\
         suitcase   & 61.61 & oven       & 40.74 & stop\_sign & 0.01 \\ \hline
         % rabbit     & 59.56 & ski        & 32.99 & crocodile  & 0.00      \\ \hline
     \end{tabular}
   \end{center}
   % \caption{Frame-level object detection mAP (\%) of our trained detector on VidOR-val.}
   \caption{Frame-level object detection mAP (\%) on VidOR-val.} 
   \vspace{-2.5em}
   \label{table_detector}
\end{table}
\addtolength{\tabcolsep}{0.5pt}

\addtolength{\tabcolsep}{-0.5pt}
% \vspace{-2em}
\begin{table}[t]
   \begin{center}
      \begin{tabular}{ccc}
         \hline
         Method             & mAP                & Position mAP   \\ \hline           
         MAGUS.Gamma~\cite{sun2019video}             & 8.82                 & 16.64               \\
         colab-BUAA~\cite{xie2020video}             & \textbf{14.59}                 & ---               \\
         Ours                    & 12.48    & \textbf{23.71}               \\ \hline
      \end{tabular}
   \end{center}
   \caption{Tracklet mAP (\%) on VidOR-val. %$(\cdot)^{\dagger}$ are evaluated by ourselves based on the released tracklets.
   } 
   \vspace{-2em}
   \label{table_tracklet_mAP}
\end{table}
\addtolength{\tabcolsep}{0.5pt}

\subsection{Datasets and Evaluation Metrics} \label{sec:datasets}
We evaluated our method on the VidOR dataset~\cite{shang2019annotating}. It consists of 10,000 videos collected from YFCC100M~\cite{thomee2016yfcc100m}, which covers 80 object categories and 50 predicate categories. We used official splits~\cite{shang2019annotating}, \ie, 7,000 videos for training, 835 videos for validation (VidOR-val), and 2,165 videos for test (VidOR-test).

We use the official evaluation metrics~\cite{shang2017video,vru} of the VRU Challenge, including Relation Detection (RelDet) and Relation Tagging (RelTag). Quantitative metrics includes Average Precision (mAP) and Recall@K (R@K, K=50,100) for RelDet, and Precision@K (P@K, K=1,5,10) for RelTag.

\subsection{Implementation Details} \label{sec:implementation} 

\noindent\textbf{Temporal Anchor Settings.} The time slots of temporal anchors (as well as the time slot of tracklets) are normalized into (0, 1) with respect to the video length. Since we assign each query with a unique temporal anchor, we have total $m$ anchors. Specifically, the $m$ anchors are associated with $m_c$ different center points and $m_d$ different durations, which are ($1/m_c,2/m_c,\ldots,1$) and ($1/m_d,2/m_d,\ldots,1$), respectively. We set $m_c$ =16, $m_d$=12. Typically, the number $m=m_cm_d$ is set to be larger than the number of most of predicates in a video.

\noindent\textbf{Parameter Settings.} 
The frame-level appearance feature dimension $d_a = 1024$. The dimension of word embdding $d_w=300$. The hidden dimension $d,~d_q$ and $d_v$ were set to 512. The output lengths of pooling operations were set as $l=4,~l_\text{roi}=7$. All the MLPs are two-layer fully-connected networks with ReLU and with hidden dimension of 512. All bounding boxe coordinates are normalized to the range between (0, 1) with respect to video size. The number of encoder/decoder layers are set as $L_e=6$ and $L_d=4$, respectively. The loss factors were set as $\lambda_\text{cls}=1.0$, and $\lambda_\text{adj}=30.0$. For training, we trained our model by Adam~\cite{kingma2014adam} with total 50 epochs. The learning rate was set to 5e-5 and the batch size was set to 4.

\noindent\textbf{Inference Details.} 
At inference time, we keep top-10 predictions for each predicate~\cite{shang2017video}. The time slot of each relation triplet is calculated as the intersection of the two corresponding tracklets. During experiments, we found that several predicate queries with same category are sometimes linked to the same tracklet pair, which causes repeated predictions in the top-K candidates. Thus, we performs a filtering operation on the final set of triplet predictions: For all triplets with the same predicate category and tracklet pair, we only keep the one with the highest score.

\subsection{Component Analysis}\label{sec:ablation}
% We run a number of ablations to analyze the influence of different hyperparameters and network designs of TempFormer. 
\noindent\textbf{Frame-level Object Detection.} Table~\ref{table_detector} presents the detection mAP (IoU=0.5) on VidOR validation set of our trained MEGA in several categories. Common categories with more samples show better performance than rare categories with few samples. The overall mAP of all categories is 41.21\%.

\noindent\textbf{Object Tracklet Detection.} 
Our tracklet proposals on VidOR validation set achieve a tracklet mAP (vIoU=0.5) of 11.67\% and an upper bound of 23.08\%, as shown in Table \ref{table_tracklet_mAP}, suppressing MAGUS.Gamma~\cite{sun2019video}. The tracklets of colab-BUAA~\cite{xie2020video} have higher mAP than ours, due to their heavier detection backbone Cascade R-CNN (C-RCNN)~\cite{cai2018cascade} with ResNeSt101 (S-101)~\cite{zhang2020resnest}.
%  compared with our MEGA with ResNet-101 (R-101).

\noindent\textbf{Relation Prediction.} 
We analyze the effects of different components (\ie, classeme features (Clsme) and predicate frequency bias (Bias)) on relation prediction, the results are shown in Table~\ref{table_val_ensemble}. It shows that the model with both Clsme and Bias achieves the highest mAP and Tagging Precision, while the model with only Bias performs slightly better on Recall in RelDet. We choose the model consisting both Clsme and Bias as our final model.

We also consider model ensembling in Table~\ref{table_val_ensemble}, where Ens-$k$ represents the results of ensembling $k$ models. Ens-5 has the highest performance in all metrics except a slightly lower mAP than Ens-3. So we use Ens-5 as our final model in the VRU Challenge.

\addtolength{\tabcolsep}{-0.5pt}
\begin{table}[t]
   \begin{center}
      \begin{tabular}{c|cc|ccc|cc}
         \hline
         \multirow{2}{*}{Model} & \multirow{2}{*}{Clsme} & \multirow{2}{*}{Bias} & \multicolumn{3}{c|}{RelDet} & \multicolumn{2}{c}{RelTag} \\
                                &                        &                       & mAP     & R@50    & R@100   & P@5          & P@10        \\ \hline
         \multirow{3}{*}{Ens-1} &                &               & 8.06    & 7.02    & 7.96    & 49.78        & 38.60       \\
                                &                & $\boldcheckmark$          & 8.36    & \textbf{7.25}    & \textbf{8.18}    & 50.73        & 39.22       \\
                                & $\boldcheckmark$           & $\boldcheckmark$          & \textbf{8.67}    & 7.10    & 8.14    & \textbf{51.49}        & \textbf{39.38}       \\ \hline
         Ens-2                  & $\boldcheckmark$           & $\boldcheckmark$          & 9.06    & 7.79    & 9.12    & 52.37        & 40.11       \\
         Ens-3                  & $\boldcheckmark$           & $\boldcheckmark$          & \textbf{9.36}    & 8.14    & 9.68    & 52.56        & 40.97       \\
         Ens-5                  & $\boldcheckmark$           & $\boldcheckmark$          &   9.33  & \textbf{8.35}    & \textbf{10.21}    & \textbf{52.73}        & \textbf{41.14}       \\ \hline
      \end{tabular}
      %9.33	8.35	10.21		52.73	41.14
   \end{center}
   \caption{Component analysis of our model on VidOR-val.} 
   \vspace{-2.5em}
   \label{table_val_ensemble}
\end{table}
\addtolength{\tabcolsep}{0.5pt}

\subsection{Comparisons with State-of-the-Arts}
We compare our model with other state-of-the-arts (SOTA) methods on VidOR test set (\ie, the VRU Challenges~\cite{ji2021vidvrd}), and the results are shown in Table~\ref{table_test_map}. We outperform all other teams, except for colab-BUAA~\cite{xie2020video}. Though colab-BUAA performs slightly better than our method, the detection backbone used in their method is C-RCNN with S-101, which is very heavy, compared to our MEGA with ResNet-101 (R-101). Besides, their features contain realative motivion feature of tracklet pairs (RM) and the location mask feature (Msk), which are also of high complexity.

We also analyze the detection backbones used in different methods in Table~\ref{table_test_map}. Different detectors have a great influence on the final relation results. In \cite{zheng2019relation}, the detector is RefineDet (R-Det)~\cite{zhang2018single} with VGG-16~\cite{simonyan2014very}, and the tracking algorithm is greedy tracklet grneration (GTG) with tracklet-NMS (T-NMS). This backbone returns tracklets with relatively poor quality. In contrast, better detectors and trackers (such as FGFA~\cite{zhu2017flow}, Seq-NMS~\cite{han2016seq}, improved Seq-NMS (iSeq-NMS)~\cite{xie2020video} and KCF~\cite{henriques2014high}) provide higher quality tracklets and improve the relation detection performance, while introducing more computational complexity.

% \noindent\textbf{VidOR Validation Set.} 

% \noindent\textbf{VidOR Test Set (VRU Challenges).} 

\section{Conclusions}
In this paper, we introduce a novel video visual relation detection method, which consists object tracklet detection and visual relation detection. Specifically, we use MEGA~\cite{chen2020memory} and deepSORT~\cite{wojke2017simple} to detect object tracklets, and we propose a tracklet-based visual Transformer for relation detection, in which a temporal-aware decoder is spatially designed. The experiment results demonstrate the superiority of our method, which outperforms other SOTA methods in the VRU Challenge of ACM Multimedia 2021.
% Video Relation Understanding Grand Challenge .
% the Video Relation Understanding Challenge.

\noindent\textbf{Acknowledgement} 
This work was supported by the National Key Research and Development Project of China (No. 2018AAA0101900), the National Natural Science Foundation of China (U19B2043, 61976185), Zhejiang Natural Science Foundation (LR19F020002), Zhejiang Innovation Foundation(2019R52002), and the Fundamental Research Funds for the Central Universities.

\bibliographystyle{ACM-Reference-Format}
\bibliography{VidSGGbib_used}

\end{document}